\documentclass[runningheads]{llncs}
\usepackage[T1]{fontenc}
\usepackage{hyperref}
\usepackage{amsmath}
\usepackage{amssymb}
\usepackage{textcomp}
\usepackage{graphicx}
\usepackage{multirow}
\usepackage{bbding}
\usepackage{threeparttable}
\usepackage{booktabs}
\usepackage{tabularx}
\usepackage{float}
\usepackage{makecell}
\usepackage{color} 

\begin{document}

\title{VS-LLM: Visual-Semantic Depression Assessment based on LLM for Drawing Projection Test}
\titlerunning{VS-LLM based on LLM for Drawing Projection Test}

\author{Meiqi Wu\inst{1} \and Yaxuan Kang\inst{2} \and Xuchen Li\inst{2} \and 
Shiyu Hu\inst{2,3} \and
Xiaotang Chen\inst{2} \and
Yunfeng Kang\inst{2} \and
Weiqiang Wang\inst{1} \and
Kaiqi Huang\inst{2,3,4}}

\authorrunning{M. Wu et al.}

\institute{}

\institute{School of Computer Science and Technology, University of Chinese Academy of Sciences,
Beijing, China \and
Institute of Automation, Chinese Academy of Sciences, Beijing, China \and
School of Artificial Intelligence, University of Chinese Academy of
Sciences, Beijing, China \and
CAS Center for Excellence in Brain Science and Intelligence Technology, Shanghai, China \\ 
\email{wumeiqi18@mails.ucas.ac.cn, \{yaxuan.kang, lixuchen2024\}@ia.ac.cn, hushiyu199510@gmail.com, xtchen@nlpr.ia.ac.cn, yunfeng.kang@ia.ac.cn wqwang@ucas.ac.cn, kqhuang@nlpr.ia.ac.cn}\\}

%
\maketitle              
\begin{abstract}

The Drawing Projection Test (DPT) is an essential tool in art therapy, allowing psychologists to assess participants' mental states through their sketches. Specifically, through sketches with the theme of ``a person picking an apple from a tree (PPAT)'', it can be revealed whether the participants are in mental states such as depression. Compared with scales, the DPT can enrich psychologists' understanding of an individual's mental state. However, the interpretation of the PPAT is laborious and depends on the experience of the psychologists. 
To address this issue, we propose an effective identification method to support psychologists in conducting a large-scale automatic DPT. 
Unlike traditional sketch recognition, DPT more focus on the overall evaluation of the sketches, such as color usage and space utilization. Moreover, PPAT imposes a time limit and prohibits verbal reminders, resulting in low drawing accuracy and a lack of detailed depiction. 
To address these challenges, we propose the following efforts: (1) Providing an experimental environment for automated analysis of PPAT sketches for depression assessment; (2) Offering a \textbf{V}isual-\textbf{S}emantic depression assessment based on \textbf{LLM} (\textbf{VS-LLM}) method; (3) Experimental results demonstrate that our method improves by 17.6\% compared to the psychologist assessment method. We anticipate that this work will contribute to the research in mental state assessment based on PPAT sketches' elements recognition. Our datasets and codes are available at: \href{https://github.com/wmeiqi/VS-LLM}.


\end{abstract}

\section{Introduction}

\begin{figure}
\vspace{-0.3cm}
\centering
\includegraphics[width=\textwidth]{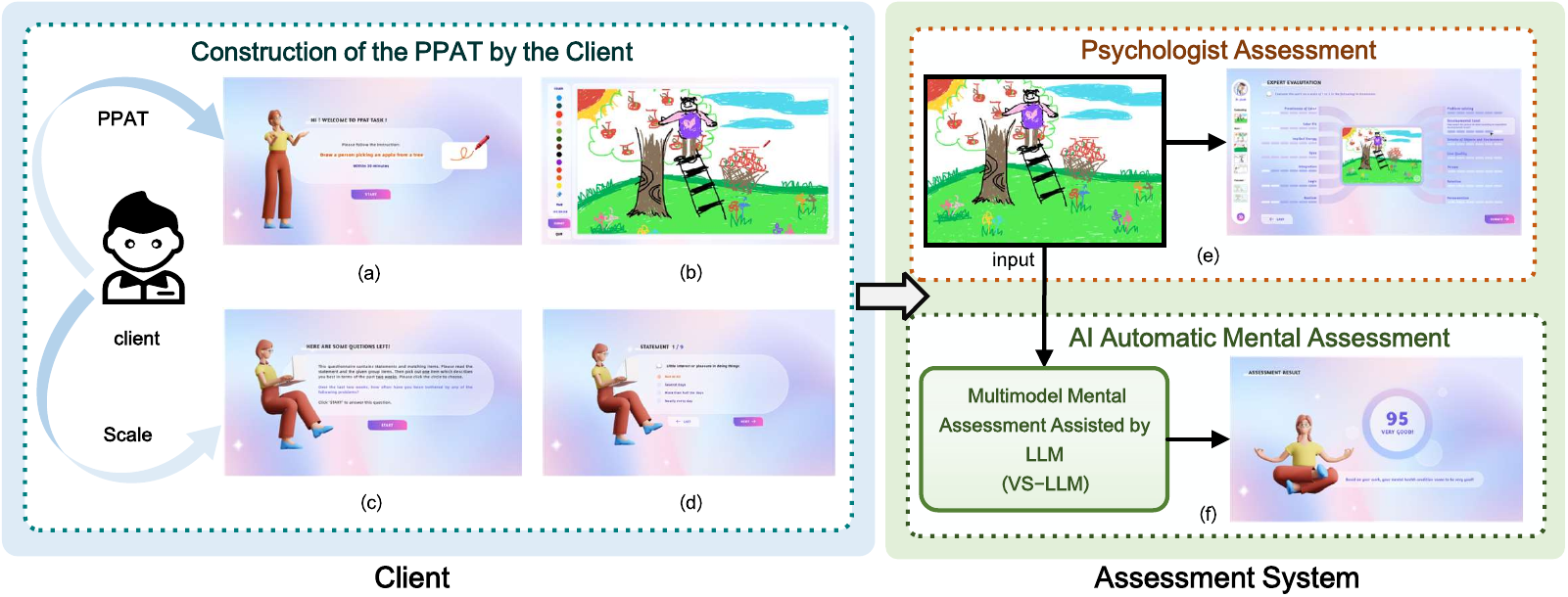}
\caption {System Flowchart. Our mental state assessment on the electronic DPT system consists of two parts. The first part involves the participant, who is asked to participate in the PPAT task and finish the scale. \textbf{(a) (b)} represents the user interface of the drawing flow of the PPAT task, while \textbf{(c) (d)} depicts the assessment flow of the participant's depression scale. The second part is the mental state assessment system. \textbf{(e)} illustrates an expert-oriented evaluation system to collect psychologists' reviews of PPAT images from 14 dimensions.  \textbf{(f)} demonstrates the automated assessment flow of VS-LLM which serves as the multi-modal psychological assessment algorithm proposed by us.
}
\label{fig:framework}
\vspace{-0.3cm}
\end{figure}

In computer vision, the sketch can be regarded as an expression of the human brain’s internal representation of the visual world~\cite{9706366}. Moreover, in the field of psychology, sketches of the Drawing Projection Test (DPT) play a crucial role in evaluating mental disorders~\cite{Au2024TheAO}. Compared with self-reported scale~\cite{kroenke2001phq}, it has been widely proven that the DPT can enrich psychologists' understanding of individuals’ mental states, such as depression, anxiety, aggressiveness, and their normal states. However, despite its potential benefits, DPT has not seen widespread adoption in clinical practice. This raises the question: \textit{Why hasn't DPT been more widely utilized in psychological assessment and treatment?}

In Fig.~\ref{fig:framework}, psychologists evaluate participants' mental states based on the ``a person picking an apple from a tree (PPAT)'' task~\cite{gantt1998formal}, which entails grading along 14 dimensions. This process demands significant time, effort, and expertise.
It is hard to deploy evaluation tasks in large-scale settings like schools or companies. Moreover, unlike traditional sketch recognition methods in computer vision~\cite{Eitz2012HowDH,yu2017sketch,he2017sketch,li2018sketch,FernandezFernandez2019QuickSA}, PPAT sketches present unique challenges due to imposed time limits and the prohibition of verbal reminders, resulting in lower drawing accuracy and less detailed depictions. To address these issues, some AI-based automatic assessment methods~\cite{Chen2020AutomaticDS,RakhmanovAA20,incollection,wu2024finger,hu2023multi,ling2025vmbenchbenchmarkperceptionalignedvideo,feng2025narrlv,li2024visual} are proposed, they primarily use CNNs to extract visual features from the final sketch. However, these approaches often overlook detailed analysis of the painting process and fail to provide comprehensive evaluations, including aspects like color usage and space utilization. In addition, there is still no unified testing environment available for study.

To address those limitations, we have developed an Artificial Intelligence (AI) automatic depression assessment method, Visual-Semantic Depression Assessment Based on Large Language Model (VS-LLM), to support psychologists in conducting a large-scale automatic DPT. In particular, the Visual Perception Module utilizes stroke information to model the painting process by decomposing the original sketch into sequences, enabling a more detailed analysis. The Mental Semantic Caption Generation Module generates mental captions by designing tailored prompts (such as color usage and space utilization) for LLM (specify what LLM stands for if possible). Subsequently, the Semantic Perception Module captures overall information under the guidance of expert knowledge. 

To verify the effectiveness of our method, we first constructed a PPAT dataset containing 690 PPAT sketches aligned with the PPAT task in psychology. Finally, the experimental results demonstrate the superior performance of our method, which improves by 17.6\% compared to the psychologist assessment method. In summary, our contributions are:
\begin{itemize}
    \item We developed a visual-semantic depression assessment system based on LLM (\textbf{VS-LLM}) method, where the visual perception module and the mental semantic caption generation module are respectively used to obtain more detailed and overall information from the painting, enabling more effective analysis of PPAT.
    \item We first provided an experimental environment for automated analysis of PPAT sketches for depression assessment.
    \item Our experiments demonstrate the superior performance of our method, which improves by 17.6\% compared to the psychologist assessment method.
\end{itemize}


\section{Related Work}

\subsection{Drawing Projection Test and PPAT}

DPT aims to depict the authentic psychological state of participants by guiding them to draw specific topics that can reflect subconscious or unconscious information that cannot be expressed in language~\cite{hammer1958clinical}. So far, it has been widely proven the effectiveness of identifying mental state in various clinical settings (e.g. depression~\cite{gu2020screening}, stress~\cite{kravits2010self}, and child sexual abuse~\cite{amir2007dissociation}). Psychologists evaluate sketches based on experiences and professional standards, such as the use of color, the detail of elements in the sketch, or the coherence and logic between elements. Silver's work reviews a body of research on the Silver Drawing Test (SDT) and Draw A Story (DAS) art-based assessments, which span 40 years of development~\cite{Silver2009IdentifyingCA}.

As a kind of DPT method, the PPAT test originated in 1990 and is currently widely used in clinical and non-clinical settings, with revisions made for different cultural backgrounds~\cite{manickam2016elements}. For example, in India, participants were asked to draw ``a person picking a mango from a tree (PPMT)'' and scored by the Formal Elements Art Therapy Scale (FEATS). It was proven that PPMT can effectively distinguish depression patients~\cite{manickam2016elements}. 
In the ``person'' and ``details and environment'' dimensions of the PPAT, Scott et al.~\cite{Gussak2006EffectsOA} discovered that patients diagnosed with depression according to DSM-5 performed significantly differently from the control group.
Numerous studies have shown that PPAT can effectively differentiate between subjects with subclinical psychological conditions such as depression, anxiety, and aggression, and those in a normal state~\cite{gantt1998formal,kruthi2023person}.

While the DPT is effective in assessing mental states, it relies on the expertise of psychologists and can be time-consuming and labor-intensive. Therefore, there is a growing trend towards developing automated detection methods.

\subsection{AI Technology for Art Therapy}
In recent years, with the rapid development of AI, AI-based art therapy has attracted the attention of many scholars. For instance, Chen et al.~\cite{Chen2020AutomaticDS} proposed a scoring system ranging from 1 to 6 based on clock drawings. By comparing the performance of VGG16~\cite{simonyan2014very}, ResNet-152~\cite{he2015deep}, and DenseNet-121~\cite{huang2018densely}, they illustrated how neural networks can effectively screen individuals for dementia and quantitatively estimate its severity. Similarly, Kim et al.~\cite{kim2024exploring} utilized ResNet for classifying mental states and numerical evaluation tasks. 

Beyond classification techniques, other researchers have explored classic detection methods. For example, Lee et al.~\cite{lee2024generating} applied deep learning to analyze House-Tree-Person sketches by extensive feature extraction and object detection, which got high accuracy in mental state analysis. Moreover, Kim et al.~\cite{kim2023alphadapr} employed YOLOv5~\cite{Jocher2021ultralyticsyolov5V} to effectively evaluate Draw-A-Person-in-the-Rain (DAPR)~\cite{handler2014drawings}. 

Compared with traditional expert assessment methods, AI-based art therapy methods will save more costs and the analysis results will be more authentic. However, they typically applied CNN to extract features from the final sketch, regardless of the details analysis of the painting process and the overall features of sketches. In addition, the sketches they analyzed were different and lacked a unified experimental environment~\cite{Hu_2023} for everyone to study.

\subsection{Large Language Model}
The rise of Large Language Models (LLMs) signifies the dawn of a transformative period in Artificial Intelligence, fundamentally restructuring the entire domain. For example, GPT-4V~\cite{openai2024gpt4} marks a significant advancement, allowing users to instruct GPT-4 to analyze image inputs. LLaVA~\cite{liu2023visual} integrates a vision encoder from CLIP~\cite{radford2021learning} with the LLM, enabling it to process visual information alongside language. QWen-VL \cite{bai2023qwen} is a multi-modal large language model that demonstrates remarkable capability in understanding and generating content that seamlessly integrates visual and linguistic elements.

Specifically, the integration of Large Language Models (LLMs) in PPAT presents innovative opportunities to enhance psychological understanding and analysis. Leveraging extensive text data for training, LLMs demonstrate superior performance in natural language processing tasks, making them as valuable tools for interpreting symbolic content in PPAT sketches. Furthermore, LLMs have the potential to automate certain aspects of PPAT image analysis, streamline workflow, and enable psychologists to focus on complex cases. 

In summary, integrating LLMs into PPAT promises to enrich psychological comprehension, offering new pathways for delving into the human psyche.


\section{Methods}

\begin{figure}
\vspace{-0.3cm}
\centering
\includegraphics[width=\textwidth]{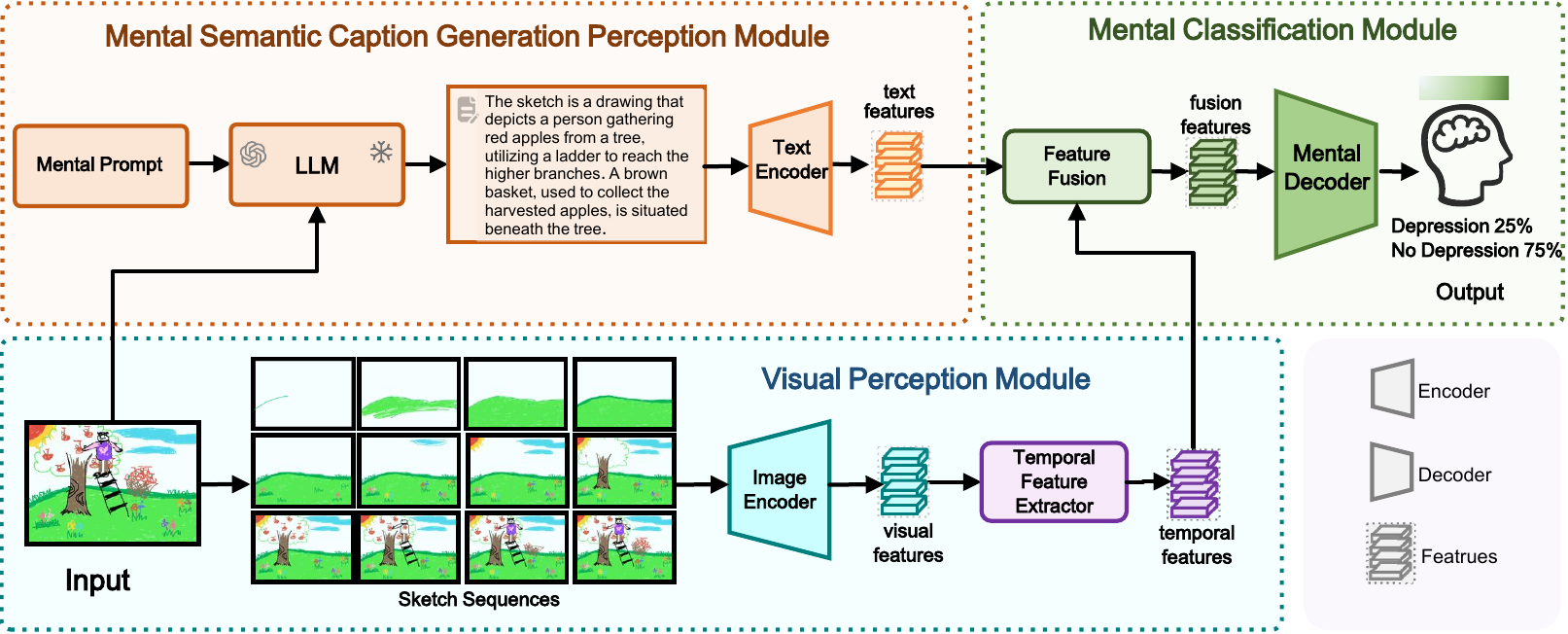}
\caption {The architecture of VS-LLM. VS-LLM comprises the Visual Perception Module, Mental Semantic Caption Generation Perception Module, and Mental Classification Module.}
\label{fig:method}
\vspace{-0.3cm}
\end{figure}

This section demonstrates how the proposed VS-LLM effectively assesses depression. As depicted in Fig.~\ref{fig:method}, VS-LLM comprises three modules. The Visual Perception Module extracts detailed visual information by decomposing the sketch into sketch sequences; The Mental Semantic Caption Generation Module generates an overall caption for the Text Encoder to obtain global information on the sketch; The Mental Classification Module integrates visual and semantic features for classification.

\subsection{Visual Perception Module}

\subsubsection{Decomposition of Sketch.}
When participants draw, they typically adhere to a specific sequence of strokes or components, providing insights into their thought processes. We propose decomposing the sketch into several sequences to capture more detailed information.

\begin{figure}
\vspace{-0.3cm}
\centering
\includegraphics[width=\textwidth]{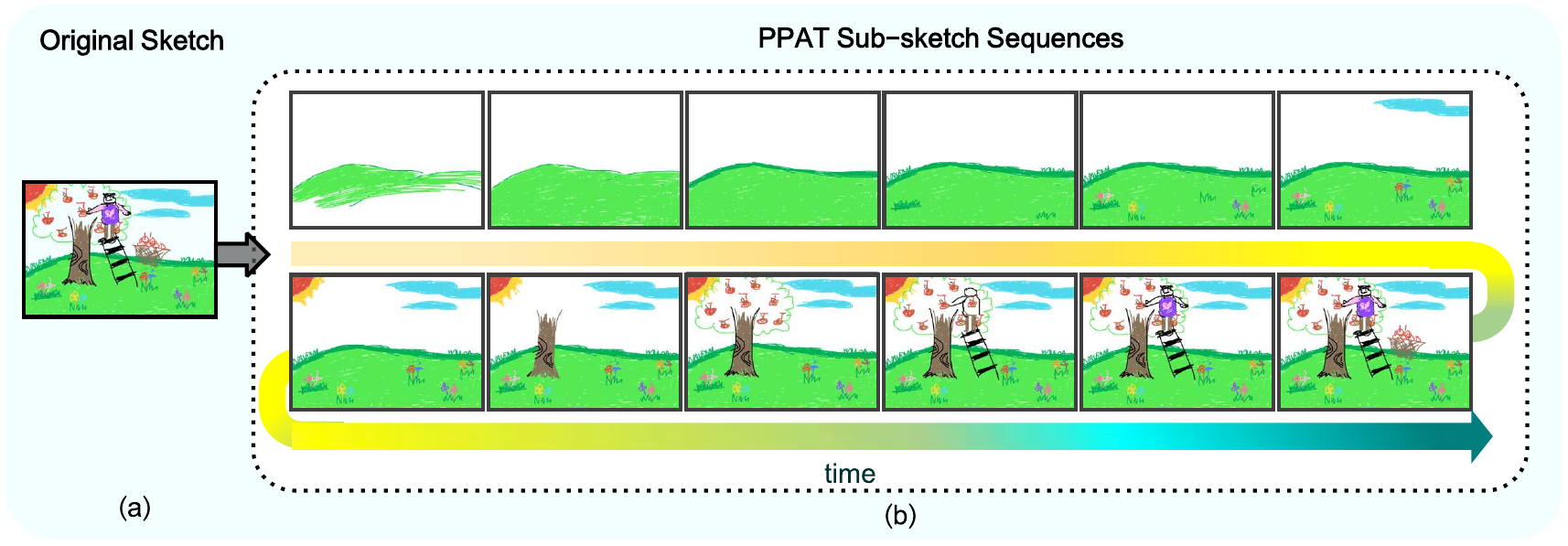}
\caption {PPAT Sub-sketch Sequence. (a) represents the complete original sketch, and (b) represents the generated sub-sketch sequence sketch. The specific order of the sub-sketch is arranged from left to right and from top to bottom, forming a sequence of sub-pictures with timing information.}
\label{fig:sub}
\vspace{-0.3cm}
\end{figure}

According to equations (\ref{eq:1}) and (\ref{eq:2}), the resulting sketch drawing is depicted in Fig.~\ref{fig:sub}. Each original sketch is decomposed into 12 consecutive sub-sketches, each containing all the stroke information of the preceding sub-sketches. For the $i$-th sketch, the division rules for sub-sketches can be formalized as:

\begin{align} \label{eq:1}
step_{i} =\frac{sn_{i} }{12},
\end{align}
\begin{align} \label{eq:2}
sn_{ij} =\begin{cases}
   j\times 1  & \left \lfloor step_{i} \right \rfloor< 1,j< sn_{i}  \\
  sn_{i} & \left \lfloor step_{i} \right \rfloor< 1,j\ge sn_{i} \ or \ \left \lfloor step_{i} \right \rfloor\ge 1,j=12\\
  j\times\left \lfloor step_{i} \right \rfloor & \left \lfloor step_{i} \right \rfloor\ge  1,j< 12
\end{cases},
\end{align}

Where $sn_i$ is the total number of strokes in the $i$-$th$ original image. $i=1,2,3,...,n$, $n$ is the number of all sketches. $step_i$ is the stroke step in the $i$-$th$ original image. $ \left \lfloor step_{i} \right \rfloor$ is $step_i$ round down. $sn_{ij}$ is the number of strokes representing the $j$-$th$ sub-sketch of the $i$-$th$ sketch, $j=1,2,...,12$.

\subsubsection{Image Encoder.}

In the Visual Perception Module, the convolution uses a part of ResNet18~\cite{he2016deep}, the input is $x\in \mathbb{R} ^{W\times H\times C} $ , where $H$,$W$ are the size of $x$ and $C$ is channels, and the output after layer-by-layer convolution pooling (removing full connections) is $(512,7,7)$. Finally add a layer of max pooling to become $(512,3,3)$~\cite{cardona2019seeing}. For sketch $x$ we have:
\begin{align}
F_{v} =MaxPooling\left (  \mathcal{\mathcal{R} } \left ( x \right )  \right )\in \mathbb{R} \mathbb{^{\mathrm{512} \times\mathrm{ 3} \times \mathrm{3} } } ,
\end{align}
where $\mathcal{R}(x)$ represents the process of ResNet processing the input sketch $x$. $F_{v}$ represents visual features.

\subsubsection{Temporal Feature Extractor.}
Temporal Feature Extractor consists of two-layer LSTM~\cite{fischer2018deep}, which captures temporal dependencies and learns semantic knowledge from high-level feature sequences. First, flatten the visual features $F_{v}$ into a one-dimensional tensor ($3\times3\times512=4,608$). Then, the flattened visual features are taken as the input of LSTM. Finally, we add a linear mapping layer.
\begin{align}
F_{t} =LSTM\left ( Flatten\left ( F_{v}  \right )  \right ) \in \mathbb{R^{\mathrm{100} \times\mathrm{12} } },
\end{align}
where $F_{t}$ represents temporal stroke features.

\subsection{Mental Semantic Caption Generation Perception Module}

\subsubsection{Mental Semantic Caption Generation.}
\begin{figure}
\vspace{-0.3cm}
\centering
\includegraphics[width=\textwidth]{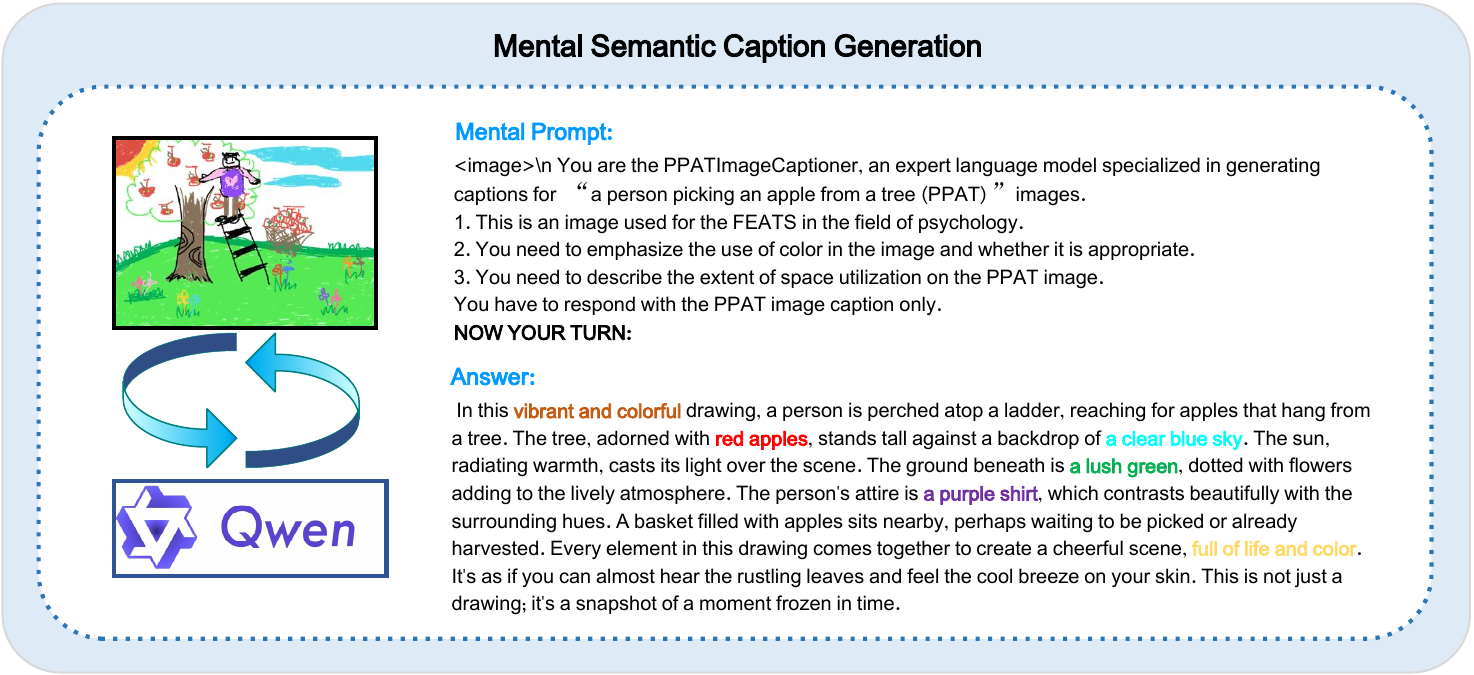}
\caption {Mental Semantic Caption Generation Framework. We designed mental prompts and input them into Qwen-VL to obtain the psychological semantic description of the sketch.}
\label{fig:MSIG}
\vspace{-0.3cm}
\end{figure}

As shown in Fig.~\ref{fig:MSIG}, to obtain advanced mental semantic captions from PPAT sketches, we designed mental prompts to utilize Qwen-VL~\cite{bai2023qwen}. These mental prompts emphasize psychological elements, such as color usage and space utilization, ensuring the quality of text generated by LLM. For PPAT sketches,
\begin{align}
    Caption_{psy} = LLM(P_{psy}, I_{ppat}),
\end{align}
where $Caption_{psy}$, $P_{psy}$, $I_{ppat}$ represent psychological captions, prompts with psychological elements, and PPAT sketches, respectively.

\subsubsection{Text Encoder.}
We use RoBERTa~\cite{liu2019roberta}, which is a transformer encoder model pre-trained on a large corpus of text data. We tokenize input sentences into a series of text tokens and then feed the token sequence into the RoBERTa language model to extract text embedding vectors. In this semantic perception module, the input text $Caption_{psy}\in \mathbb{R} ^{W \times C} $, where $W$ is the length of the text sequence and $C$ is the dimension of the features, while the output is the feature representation transformed through various layers. 
\begin{align}
F_{t} = \text{Pooling}\left( \mathcal{RO}(Caption_{psy} ) \right) \in \mathbb{R}^{\mathrm{D}},
\end{align}
Where $\mathcal{RO}(Caption_{psy})$ represents the process of RoBERTa model processing the input text $Caption_{psy}$, $Pooling$ represents global pooling operation, and $F_t$ is the final text feature representation with dimension $D$.

\subsection{Mental Classification Module}
After extracting semantic and visual information, we concatenate them. Then, we design a mental decoder. Specifically, the mental decoder consists of three consecutive linear layers to further learn multi-modal mental features. The decoder outputs psychological categories.

\subsection{Training with Focal Loss}
To address the imbalance between positive and negative samples, we utilized Focal Loss during the training process ~\cite{lin2017focal}. The abundance of negative samples, constituting the majority of the total loss, often consists of easily classifiable instances. Consequently, the model's optimization direction may not align with our expectations. This function aims to redirect the model's focus towards hard-to-classify samples by attenuating the weight of easy-to-classify ones.
Focal Loss introduces a modulating factor $(1 - p_{t})^\gamma$ to the cross-entropy loss, where $\gamma \ge 0$ is a tunable focusing parameter. We define the focal loss as:
\begin{align}
 {FL(p_{t}) = - (1 - p_{t})^\gamma \log (p_{t})}.
\end{align}

The focal loss is visualized for several values of $\gamma \in [0,5]$. 
The focusing parameter $\gamma$ smoothly adjusts the rate at which easy examples are down-weighted. When $\gamma = 0$, Focal Loss is equivalent to Cross-Entropy Loss (CE), and as $\gamma$ increases, the effect of the modulating factor becomes more pronounced.

\label{sect:figures}

\section{Dataset Contruction}
To establish an experimental environment for the automated analysis of PPAT sketches, we constructed a PPAT dataset. 
This section begins by discussing the collection and annotation process of the dataset, followed by the presentation of statistics and division of the dataset.

\subsection{Dataset Collection and Annotation}

\begin{table}
\centering
\caption{PPAT Dataset Statics.}
\label{table:statics}
\renewcommand{\arraystretch}{1.5}
\begin{tabular}{ccc}
\toprule
\multicolumn{1}{c}{\textbf{Metric}}      & \textbf{Type}             & \textbf{Value}   \\ \toprule
\multicolumn{1}{c}{\multirow{4}{*}{Age}} & 0-20             & 123/690 \\
\multicolumn{1}{c}{}                     & 20-40            & 531/690 \\
\multicolumn{1}{c}{}                     & 40-60            & 32/690  \\
\multicolumn{1}{c}{}                     & \textgreater{}60 & 4/690   \\ \hline
\multirow{3}{*}{Gender}                  & Male             & 302/690 \\
                                         & Female           & 388/690 \\
                                         & Neutral          & 0/690   \\ \hline
\multirow{2}{*}{Mental State}                  & Depression       & 117/690 \\
                                         & No Depression    & 573/690 \\ \bottomrule
\end{tabular}
\end{table}

\subsubsection{Dataset Collection Process.}
As depicted in Fig.~\ref{fig:framework}, participants were invited to participate in the research project. The software operation process involved providing basic personal information and completing sketch tasks, which consisted of two parts: sketch adaptation exercises and formal sketch tasks. Additionally, participants were required to complete the Patient Health Questionnaire (PHQ-9) scale~\cite{kroenke2001phq}, which is an established diagnostic instrument designed to screen for the presence and measure the severity of depressive symptoms in individuals.

\subsubsection{Dataset Annotation.} The sketch was evaluated by three psychologists according to FEATS, and the average score was calculated. Subsequently, each sketch obtained 14 scores, corresponding to the 14 dimensions in FEATS, thus forming a 14-dimensional vector of real numbers. The ground truth is the results of the PHQ-9 scale.

\subsubsection{Checkout Flow.} 
We have implemented a stringent data review process to uphold the dataset's quality. Trained professional participants understand the nuances of the PPAT task. Subsequently, the psychologist conducted a second-round review of the collected data. Finally, authors and psychologists make the ultimate judgment of whether to accept or reject the data in the third-round confirmation. Any rejection during self-check, verification, or data acceptance necessitates recollection. We believe this verification mechanism ensures the generation of a high-quality dataset.

\subsection{Dataset Statistics and Division}

We have used a threshold of 10 points for PHQ-9 scores and the statistics of depression and non-depression are provided in Table~\ref{table:statics}. In addition, we also collected the gender and age information for the dataset. As shown in Table~\ref{table:dataset_div}, we split the train dataset and test dataset by using 5-fold cross-validation for conducting experiments.

\begin{table}[htbp]
\centering
\renewcommand{\arraystretch}{1.5}
\vspace{-0.3cm}
\caption{The division of PPAT Dataset.}
\label{table:dataset_div}
\begin{tabular}{cccc}

\toprule
\makebox[0.1\textwidth][c]{}      & \makebox[0.2\textwidth][c]{\textbf{Depression}} & \makebox[0.2\textwidth][c]{\textbf{No Depression}} & \makebox[0.2\textwidth][c]{\textbf{Total}} \\ \toprule
Train & 88         & 465           & 553   \\ 
Test  & 29         & 108           & 137   \\ \midrule
Total & 117       & 573           & 690  \\ \bottomrule
\end{tabular}
\vspace{-0.3cm}
\end{table}

\section{Experiment}
In this section, we first introduce the experimental implementation details. Secondly, we conduct experiments to compare the performance of  the psychologist assessment and the AI automatic assessment. Thirdly, we demonstrate the effectiveness of different modules and Focal Loss through ablation studies.

\subsection{Implementation Details}
The VS-LLM is constructed by the PyTorch\cite{paszke2017automatic}, and the Focal Loss is implemented according to~\cite{lin2017focal}. The network is optimized with the Adam~\cite{kingma2014adam} by setting the initial learning rate to 0.001 and the batch size to 8. All the experiments are performed on a workstation with an NVIDIA GTX Titan X 12G GPU.

\begin{table}
\vspace{-0.3cm}
\centering
\renewcommand{\arraystretch}{1.5}
\caption{Experimental Results of Psychologist Assessment and AI-Automated 
Assessment on PPAT Dataset. The input for the psychologist assessment method consists of scores from the 14 dimensions of FEATS. The input for the AI automated assessment method is the PPAT sketch. Note that we only calculate the number of trainable parameters of the model.}
\label{table:exp}
\resizebox{\textwidth}{!}{
\begin{tabular}{llccc}
\hline
\multicolumn{2}{c}{Method}                                 & \makebox[0.25\textwidth][c]{Acc(\%)}         & \makebox[0.2\textwidth][c]{Params(M)}            & \makebox[0.2\textwidth][c]{FLOPs} \\ \hline 

\multirow{4}{*}{Psychologist Assessment}                    &\makebox[0.2\textwidth][l]{Random Forest~\cite{rigatti2017random}}         & \textbf{70.2}                                    &   -   &   -  \\
                                   & \makebox[0.2\textwidth][l]{SVM~\cite{cherkassky2004practical}}            & 57.3    &     -      &    -   \\
                               & \makebox[0.2\textwidth][l]{Logistic Regression~\cite{wright1995logistic}} & 56.4    &     -      &    -     \\
                              & \makebox[0.2\textwidth][l]{MLP~\cite{hu2010pattern}}               & 64.3    &     -      &    -   \\ \hline
 \multirow{3}{*}{AI Automatic Assessment}  & \makebox[0.2\textwidth][l]{Resnet18~\cite{he2015deep}}    & 83.3     &   11.18   &    4.11   \\
           & \makebox[0.2\textwidth][l]{Sketch-a-net~\cite{yu2017sketch}}  &  85.7    &  8.41  &    1.51   \\ \cline{2-5} 
         & \makebox[0.2\textwidth][l]{\textbf{VS-LLM (Ours)}}      & \textbf{87.8}    &  8.87  &    1.51   \\ \bottomrule
\end{tabular}}
\vspace{-0.3cm}
\end{table}

\subsection{The Performance of Psychologist Assessment}
When psychologists psychoanalyze participants based on sketches, they quantitatively analyze the sketches' scores on 14 global variables (color prominence, appropriate color, effort, space use, integration, authenticity, logic, problem-solving, development level, detail and environment, line quality, people, rotation, continuous repetition). These variables are scored on a scale of 0 to 5.

Based on psychologist scores, we compared the use of Random Forest~\cite{rigatti2017random}, SVM~\cite{cherkassky2004practical}, Logistic Regression~\cite{wright1995logistic}, and MLP~\cite{hu2010pattern} to analyze the depression of the visitors. The experimental results are given in the Psychologist Assessment Method of Table~\ref{table:exp}. Among those methods, the random forest approach shows the best results for psychologist ratings, with only 70.2\% prediction accuracy on the test.

\subsection{The Performance of AI Automatic Assessment}

Regarding AI automated assessment, we conducted experiments, as shown in Table~\ref{table:exp}, titled ``AI Automatic Assessment''. Compared to the psychologist assessment, AI automated assessment outperform those of psychologists. We attribute this result to two main factors: firstly, psychologist ratings are heavily influenced by the skills and cognitive levels, and each individual may have biases in their interpretation of sketches. On the other hand, the psychologist assessment relies on scores across 14 dimensions as input, which is relatively simplistic, whereas AI automated assessment utilizes sketches as input, allowing for the extraction of richer visual features. This greatly aids in the assessment of depression.

\begin{figure}
\vspace{-0.3cm}
\centering
\includegraphics[width=\textwidth]{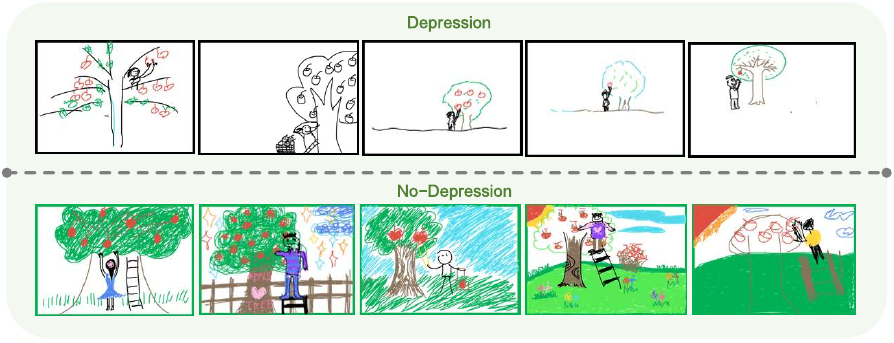}
\caption {Visualization of Recognition Results. \textbf{Top:} the examples of depression; \textbf{Bottom:} the examples of no-depression.}
\label{fig:result}
\vspace{-0.3cm}
\end{figure}

In AI automated assessment methods, our VS-LLM method have performed the best on the PPAT dataset, achieving an accuracy of 87.8\%. This success stems from the integration of psychological semantic descriptions, assisted by LLM, into the VS-LLM framework. The visualization of recognition results is shown in Fig.~\ref{fig:result}, revealing that in examples of depression, the colors appear dull and there is low utilization of space. 

Compared with the best Psychologist Assessment method, our method improves by 17.6\%, indicating that our method surpasses the level determined by psychologist ratings. This suggests the potential for automated depression assessment.

\subsection{Ablation Study}
Table~\ref{table:astudy} presents the results of ablation studies. Specifically, the experimental results in the first and 5th rows demonstrate the effectiveness of the Mental Semantic Caption Generation Module, with a performance improvement of 3.7\% achieved by incorporating psychological semantic descriptions. Comparing the results between the second and 5th rows highlights the effectiveness of the Temporal Feature Extractor, which achieved an improvement of 4.5\%. The third and 5th rows demonstrate the performance of different image encoders, with our method achieving an improvement of 0.6\%. Finally, contrasting the 4th and 5th rows, we compared the impact of different loss functions on performance, with focal loss (FL) outperforming cross-entropy loss (CEL) by 1.4\%.

\begin{table}
\vspace{-0.3cm}
\centering
\large
\renewcommand{\arraystretch}{1.5}
\caption{Ablation Studies on PPAT Dataset. FL is the abbreviation of focal loss and CEL is the abbreviation of cross-entropy loss. }
\label{table:astudy}
\resizebox{\textwidth}{!}{
\begin{tabular}{cccccc}
\toprule

 \makebox[0.2\textwidth][c]{\multirow{2}{*}{Number}} &    \makebox[0.3\textwidth][c]{\multirow{2}{*}{\makecell[c]{Mental Sematic \\ Caption Generation}}}   &  \multicolumn{2}{c}{Visual Perception Module} 
& \makebox[0.15\textwidth][c]{\multirow{2}{*}{Loss}} &  \makebox[0.15\textwidth][c]{\multirow{2}{*}{Acc(\%)}}\\ \cline{3-4}

           &                 &  \makebox[0.25\textwidth][c]{Image Encoder}  &  \makebox[0.4\textwidth][c]{Temporal Feature Extractor} &  &     \\ \toprule
1          & \XSolidBrush      & Resnet18       & \Checkmark      & FL           & 84.1                  \\
2          & \Checkmark       & Resnet18       & \XSolidBrush    & FL           & 83.3                      \\
3          & \Checkmark         & Sketch-a-Net   & \Checkmark      & FL           & 87.1                        \\
4          & \Checkmark         & Resnet18       & \Checkmark      & CEL          &  86.4                   \\ \bottomrule
\textbf{5(Ours)}    & \Checkmark        & Resnet18      & \Checkmark    & FL    & 87.8                   \\ \bottomrule
\end{tabular}
}
\vspace{-0.3cm}
\end{table}

\section{Conclusion, Discussion, and Future Work}

In this paper, We developed a visual-semantic depression assessment based on LLM (\textbf{VS-LLM}), where the visual perception module and the mental semantic caption generation module are respectively used to obtain more detailed and overall information from the sketch, enabling more effective analysis of PPAT. We provided an experimental environment for automated analysis of PPAT sketches for depression assessment. Our experiments demonstrate the superior performance of our method and confirm the importance of incorporating mental descriptions assisted by LLM. 

We have outlined the main challenges and provided insightful perspectives on automated psychological analysis. Moving forward, we will further explore anxiety assessment and other related aspects. Moreover, there are numerous promising opportunities for creating a platform that enhances research and assessment for both researchers and patients.

\section*{Acknowledgements}

This research was supported by the National Natural Science Foundation of China Grant No.62176255.

%
%
%

\end{document}